\documentclass[runningheads]{llncs}
\usepackage[T1]{fontenc}
\usepackage{graphicx}
\usepackage{booktabs}
\usepackage{xcolor}
\usepackage{algorithm}
\usepackage{algpseudocode}
\usepackage{amsmath}
\usepackage{amssymb}
\usepackage{dsfont}
\usepackage{cleveref}
\crefname{section}{section}{sections}
\usepackage[misc]{ifsym}
\usepackage{appendix}
\usepackage{caption}
\usepackage{subcaption}

\newcommand{\corr}{(\Letter)}
\newcommand{\sparsity}{\texttt{Zero Inflation}}
\newcommand{\densification}{\texttt{Zero Imputation}}
\newcommand{\remove}{\texttt{Feature Removal}}
\begin{document}

\title{Are Tabular Foundation Models Robust to Realistic Query Distribution Shifts\ \\ in Microbiome Data?}

\toctitle{Are Tabular Foundation Models Robust to Realistic Query Distribution Shifts in Microbiome Data?}



\tocauthor{Giulia Perciballi, Ahmad Fall, Federica Granese, Jean-Daniel Zucker, Edi Prifti}

\author{
Giulia Perciballi\corr\inst{1} \and
Ahmad Fall\inst{1} \and
Federica Granese\inst{2} \and
Jean-Daniel Zucker\inst{1,3} \and
Edi Prifti\inst{1,3}
}
\authorrunning{G. Perciballi et al.}

 \institute{
 IRD, Sorbonne Université, Unité de Modélisation Mathématique et Informatique des Systèmes Complexes (UMMISCO), Paris, France
 \email{\{giulia.perciballi, ahmad.fall, jean-daniel.zucker, edi.prifti\}@ird.fr}
 \and
 Inria, CNRS, I3S, Université Côte d’Azur, Valbonne, France
\email{federica.granese@inria.fr}
 \and
 INSERM, Nutrition et Obésités; Approches Systémiques, NutriOmique, AP-HP, Hôpital Pitié-Salpêtrière, Paris, France
 }

\maketitle              

\begin{abstract}
Tabular foundation models (TFMs) achieve strong performance on microbiome abundance data, yet their robustness under realistic distribution shift remains poorly characterized. We introduce a benchmark that evaluates the robustness of TFMs to biologically inspired perturbations across six gut microbiome datasets spanning four disease contexts. In this in-context learning setting, models receive unperturbed support sets as context and are evaluated on perturbed query samples. To isolate robustness beyond "shortcut" features, we preserve the most discriminative taxa and apply three controlled perturbation strategies: (i) removal of high-abundance (uninformative) taxa, (ii) sparsification via increased zero-inflation, and (iii) zero-imputation via spurious non-zero injections. 
Our results show that protecting discriminative features is insufficient to guarantee stability under support-query shift: across datasets, all perturbations degrade models' performances, with zero-imputation consistently the most harmful, indicating that corrupting global feature structure can break generalization even when key taxa are retained. Sparsification disproportionately affects TFMs relative to a classical random-forest baseline, suggesting greater sensitivity to zero-inflation–type shifts. We provide the code publicly available at \texttt{\url{https://github.com/UMMISCO/metagenomics-fm/}}.

\keywords{tabular foundation models \and robustness \and metagenomics \and zero-inflation \and distribution shift \and query-time perturbations}
\end{abstract}

\section{Introduction}
The human microbiome, meaning the collection of microorganisms residing in and on the human body, plays a key role in influencing, directly or indirectly, a broad range of physiological processes in both health and disease. Consequently, microbiome datasets have become an important testing ground for machine learning methods in biomedical applications.
Microbiome studies characterize microbial communities by profiling the relative abundance of taxa across samples, producing matrices in which rows correspond to samples and columns to taxa (i.e., features). Compared to standard tabular data, microbiome data exhibit two distinctive structural properties: \textit{compositionality}, since features represent relative abundances that sum to a constant, and pronounced \textit{zero-inflation}, as many taxa are unobserved in a given sample due to biological absence or limited sequencing depth~\cite{busato2023compositionality}. In addition, substantial variability arises across cohorts due to differences in populations, sequencing protocols, and preprocessing pipelines.

Particularly, in this work, we focus on the gut microbiome, which has been associated with a wide range of systemic diseases, including type 2 diabetes, obesity, inflammatory bowel disease, and liver cirrhosis, and it is  known to be particularly diverse compared to other body 
sites, and its composition varies considerably among healthy 
individuals~\cite{shreiner2015gut}. These sources of variability make such datasets a challenging setting for machine learning models, 
particularly when models trained or conditioned on one cohort are 
applied to samples drawn from different populations or experimental 
conditions~\cite{busato2023compositionality}.

Tabular foundation models (TFMs) have emerged as a powerful alternative to classical machine learning approaches in low-data regimes, such as those encountered in microbiome data. These models leverage in-context learning (ICL): at inference time, a labeled support set is provided as context, and predictions are generated in a single forward pass without parameter updates. Because ICL relies on similarity between support and query examples rather than parameter updates, its effectiveness may be particularly sensitive to distributional mismatch between the contextual examples and the query sample. Despite their strong performance on standard tabular benchmarks~\cite{grinsztajn2025tabpfn}, it is unclear if they are well-suited to the compositional and sparse structure of microbiome data. To investigate this issue, we provide the following contributions:

\begin{enumerate}
\item We present an evaluation protocol tailored to TFMs based on \textit{query distribution shift}, in which the support set remains unperturbed while perturbations are applied only to query samples. This design isolates the robustness of in-context learning under support–query mismatch (cf.~\Cref{sec:evaluation}).
\item We develop three biologically inspired perturbation algorithms to modify key structural properties of microbiome abundance data, namely compositionality and zero-inflation, while preserving informative features. This design isolates model robustness to distributional shifts without removing the underlying predictive signal (cf.~\Cref{sec:perturbations}).
\item Through extensive experiments on six gut microbiome datasets and six state-of-the-art TFMs, we analyze how different forms of distributional perturbations affect tabular foundation models compared to classical baselines, revealing distinct robustness patterns across perturbation types (cf.~\Cref{sec:results}).
\end{enumerate}

\subsection{Related works}
\label{sec:related}
Traditional machine learning models such as XGBoost~\cite{chen2016xgboost}, Random Forests~\cite{breiman2001random} and multi-layer perceptrons have long dominated tabular data classification, but they require task-specific training and rely heavily on hyperparameters and encoded inductive biases. To better capture the complex patterns in tabular data, recent approaches have reframed tabular prediction as a table completion problem, where missing entries are predicted from observed ones. From this perspective, TabPFN~\cite{hollmann2022tabpfn} was introduced as the first foundation model for tabular data, leveraging ICL to classify new examples directly from prompts without gradient-based fine-tuning.
Building on TabPFN, several models have extended this paradigm to larger datasets and higher-dimensional feature spaces. TabICL~\cite{qu2026tabiclv2} uses a factorized transformer with column-wise and row-wise attention to scale ICL efficiently. TabDPT~\cite{ma2024tabdpt} combines retrieval-augmented transformers with Flash Attention to handle real-world tabular datasets without dataset-specific tuning. ContextTab~\cite{spinaci2025contexttab} incorporates semantic embeddings of column names to leverage feature meanings in large-scale tables. These models demonstrate how the table-completion framing enables transformer-based foundation models to generalize across diverse tabular tasks while avoiding per-task retraining.
Several extensions have been built on this idea. TabPFN-Wide~\cite{kolberg2025tabpfn} and TabFLEX~\cite{zeng2025tabflex} apply ICL to wider and more complex tables, while newer versions of TabICL and TabPFN (\Cref{subsec:versions}) further improve efficiency and feature coverage through factorized transformer architectures and optimized prompts. Among domain-specific applications, the authors in~\cite{perciballi2024adapting} have, instead of modifying the architecture of TabPFN, adapted the prior of PFN models to generate metagenomic-like synthetic data.

\section{Preliminaries}
\label{sec:preliminaries}


\subsection{Microbiome abundance data}
Microbiome abundance data is represented as a matrix (a taxonomic abundance table) $\mathbf{X} \in \mathbb{R}^{n \times d}$, where rows 
represent samples, columns represent taxa (i.e., the features from a machine learning perspective), and each entry $\mathbf{X}_{ij}$ denotes 
the abundance of taxon $j$ in sample $i$. This data type exhibits two 
structural properties that distinguish it from generic tabular data~\cite{busato2023compositionality}:
\begin{enumerate}
    \item \textbf{Compositionality}. In taxonomic abundance tables, the individual components of a sample sum to a constant, and consequently, the features lie on a simplex, which means increasing the abundance of one taxon necessarily decreases the relative contribution of others.
    Formally, $\mathbf{X}$ is a compositional set if and only if $\forall i\in\{1, \dots, n\}$ the vector row $\mathbf{x}_i$ of $\mathbf{X}$ is in the simplex $$\mathcal{C}^{d} = \left\{\mathbf{x}_i\in\mathbb{R}^{d}\,:\,\forall j, x_{ij} > 0; \sum_{j=1}^d x_{ij} =\kappa\right\},$$ where $\kappa > 0$ is a constant, generally 1.
    \item \textbf{Sparsity}.  
    Taxonomic abundance tables have a high percentage of zero values, implying right-skewed distributions with considerable point mass at zero (\textit{zero-inflation}). Notably, zeros should be treated with caution, as they may result either from the biological heterogeneity of microbial communities, in which many organisms are detected in only a few samples, or from technical limitations that prevent the detection of some low-abundance taxa.  
\end{enumerate}
The two above properties directly motivate the 
algorithms in~\Cref{sec:perturbations}. Compositionality implies that 
locally perturbing one component, i.e., a taxon, directly impacts all 
other components of the sample, motivating our Algorithm 1) and 
Algorithm 3). The zero-inflation 
structure makes models sensitive to perturbations that alter the 
zero/non-zero pattern of the samples, motivating our Algorithm 2).

\subsection{Tabular foundation models}
TFMs are a new class of neural networks tailored 
for tabular data. Their outstanding performance stems from performing predictions via 
\textit{ICL}: at inference time, a labeled 
\textit{support set} $\mathcal{S}$ is provided as context, and the model 
produces predictions for unseen \textit{query samples} $\mathcal{Q}$ in a 
single forward pass, without any parameter updates. TFMs are based on the Prior-Data Fitted Networks (PFNs) 
framework~\cite{muller2021transformers}, in which a transformer is pre-trained 
on synthetic datasets sampled from a prior over tasks. This process meta-learns 
an approximation to Bayesian posterior predictive inference, allowing the model 
to generalize to new tasks at test time, provided the target distribution is 
consistent with the pretraining prior.

TFMs have shown strong performance across a wide range of use cases. 
However, their robustness on microbiome abundance data remains poorly 
understood. While a previous work~\cite{perciballi2024adapting} explored adapting the prior distribution of TabPFN~\cite{hollmann2022tabpfn} to this data type, it did not 
systematically analyze the failure modes of TFMs in this domain. 
Furthermore, that work preceded the recent proliferation of TFMs, and 
more capable models are now available. This motivates our systematic robustness evaluation of TFMs on microbiome data.

\subsection{TFM pretraining} The pretraining regime fundamentally shapes TFMs' behavior at inference time. TabPFN and TabICL are both pretrained on synthetic datasets sampled from a structured prior over tasks, and consequently carry no implicit prior over real feature distributions. However, TabICL introduces a column-wise Set Transformer that is explicitly distribution-aware with respect to the support set. ConTextTab and TabDPT, on the contrary,  utilize large-scale pretraining on real-world data, T4 \cite{gardner2024large} and OpenML \cite{vanschoren2014openml} datasets, respectively. As a result, they develop priors over realistic data distributions.




\begin{figure}[t]
    \centering
    \includegraphics[width=0.9\linewidth]{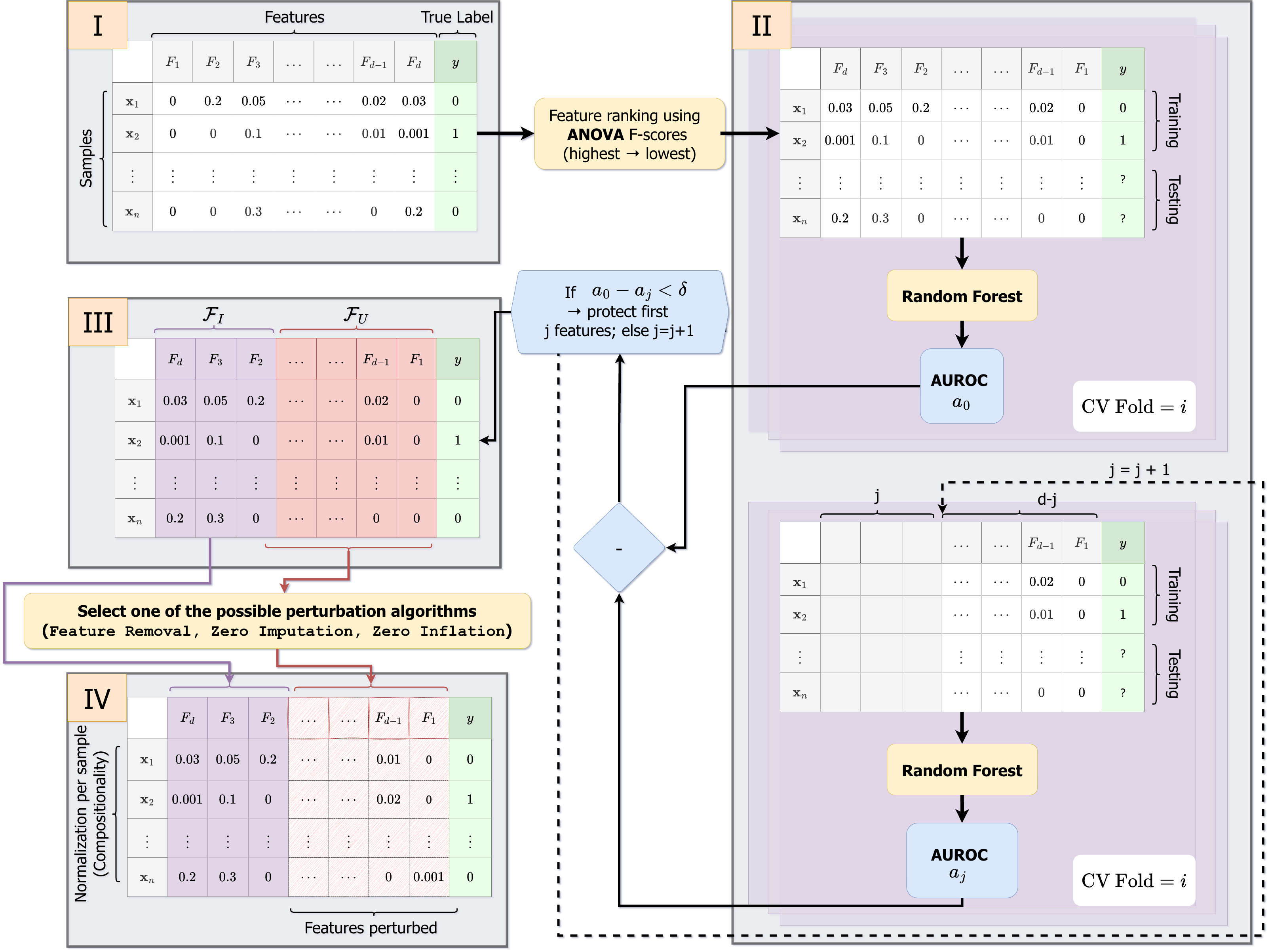}
    \caption{\textbf{Perturbation pipeline} -- Starting from the raw taxonomic abundance 
matrix $\mathbf{X}$, informative features $\mathcal{F}_I$ are identified 
via ANOVA F-test and Random Forest and protected from perturbation. One of three perturbation 
algorithms is then applied exclusively 
to the uninformative features $\mathcal{F}_U$. The perturbed matrix 
$\mathbf{X}^\prime$ is finally reconstructed by concatenating 
$\mathcal{F}_U^\prime$ with the original $\mathcal{F}_I$, followed by 
row-wise renormalization to preserve the compositional constraint.}
    \label{fig:perturbation_pipeline}
\end{figure}

\section{Method}
\label{sec:method}
This section describes the proposed framework to assess the robustness of 
TFMs on microbiome abundance data. We first introduce our biologically inspired perturbation strategies, 
and then we describe the evaluation protocol. Additional details on the 
experimental parameters are provided in~\Cref{sec:exp_setting}.
\subsection{Perturbation pipeline}
\label{sec:perturbations}
The goal of the perturbation pipeline (cf.~\Cref{fig:perturbation_pipeline}) is to design a framework introducing controlled distribution shifts targeting the two structural properties of microbiome data: compositionality and sparsity 
(i.e., zero-inflation).
In what follows, we will call $\mathbf{X}^\prime$ the taxonomic abundance matrix after perturbations. Notably, in our algorithms, we require that $\mathbf{X}^\prime$ remains compositional. 
We relegate in Section A.2 (of supplementary materials) the pseudocode of the proposed perturbation algorithms. We provide the mean abundance distributions of individual features across increasing perturbation levels, stratified by class in Figure 5 (of supplementary materials).

\subsubsection{\textit{Informative} vs.\ \textit{uninformative} feature selection.}
To avoid trivial failure modes, we first distinguish \textit{informative} 
from \textit{uninformative} features. Specifically, we define a feature as \textit{informative} if it contributes significantly to the predictive performance of a model on the task of interest. We note that features deemed important by a machine learning model may not coincide with those considered relevant by a domain expert. However, since our goal is to assess model robustness to query distribution shift, we adopt a data-driven notion of feature importance. Perturbing highly discriminative taxa would trivially degrade performance and therefore provide limited insight into robustness under distribution shift. Our perturbations are therefore applied exclusively to 
uninformative features, isolating robustness beyond reliance on shortcuts.

Let $\mathbf{y} \in \mathcal{Y}^n$ denote the vector of target 
labels for the task of interest. We first rank features by importance via 
one-way ANOVA F-tests, which provides a model-agnostic ordering without 
requiring a secondary learner. To determine the number of truly informative 
features in a dataset, we apply the following procedure. 
A Random Forest classifier is trained on the full dataset under cross-validation, 
yielding a baseline AUROC $a_0$. We then iteratively remove the top-ranked 
feature and recompute the AUROC $a_j$ after each removal step $j$. As long 
as $a_{j-1} - a_j < \delta$ (i.e., the drop is below threshold $\delta = 0.03$), 
the removed feature is considered informative. The process stops at the first 
step $j^\star$ where the drop exceeds $\delta$, and we define 
$\mathcal{F}_I = \{1, \dots, j^\star\}$ as the set of informative features 
and $\mathcal{F}_U = \{j^\star + 1, \dots, d\}$ as the uninformative ones.
Indeed, microbiome abundance data are known for their high heterogeneity 
across datasets~\cite{busato2023compositionality}. Fixing the number of 
informative features a priori (e.g., always retaining the top-5 ranked 
features) would introduce evaluation bias: performance differences across 
datasets could then reflect the arbitrary choice of cutoff rather than 
true differences in model robustness (cf. Figure 6 of supplementary materials). 
By keeping $\mathcal{F}_I$ fixed, 
perturbations target only uninformative features, ensuring that performance 
differences reflect the model's sensitivity to distributional shift rather 
than loss of essential signal.

\subsubsection{Perturbations algorithms}
Given the set $\mathcal{F}_{U}$ and the taxonomic abundance matrix $\mathbf{X}$, we consider the following perturbation strategies.

\paragraph{{\remove} (\textit{Algorithm 1}.)}

It evaluates model robustness to changes in feature availability, a common reason of variability in microbiome data, where differences in sequencing protocols can result in certain taxa being observed in some samples but absent in others~\cite{forry2024variability}. We remove the $k$ uninformative features with the highest mean abundance across samples. This creates a challenging scenario because, although uninformative for classification, high-abundance taxa contribute substantially to the data's compositional structure. The renormalization of the rows after removal of such taxa systematically inflates the relative abundances of the remaining ones, shifting their marginal distributions and altering the global compositional structure. This perturbation strategy most directly 
stresses the compositional structure of microbiome data.

\paragraph{{\sparsity} (\textit{Algorithm 2}.)}
It directly targets the zero-inflation structure of 
microbiome data, simulating the variability in sparsity patterns that arises from 
technical differences in how microbiome samples are processed and sequenced. 
Given a target sparsity level $\rho^{\star}$, we compute the number of zeros $k$ to introduce into $\mathcal{F}_U$ as $k = \min\{\rho^\star - \rho,\, \phi_U\}$, where $\rho$ is the current number of zeros, $\phi_U$ is the number of non-zero entries available in $\mathcal{F}_{U}$. We then apply power transformation
\begin{align}
\label{eq:algo_2}
    \tilde{x}_{ij} = (x_{ij})^{1+\gamma}\cdot
    \mathds{1}\left[\,(x_{ij})^{1+\gamma} \ge \tau\right],
    \qquad j \in \mathcal{F}_U,
\end{align}
where $\gamma > 0$ controls the degree of shrinkage and $\tau = 10^{-6}$ 
is a fixed detection threshold.~\Cref{eq:algo_2} preferentially 
suppresses low-abundance taxa while preserving the relative ordering 
among dominant features. To reach the target zero count $k$, we exploit 
the fact that larger values of $\gamma$ produce more aggressive shrinkage 
and therefore more zeros. We search for the $\gamma$ that minimizes 
$|\rho_U^\prime - k|$ via binary search, where $\rho_U^\prime$ counts the newly introduced zeros at each iteration. The best iterate is 
retained if exact convergence is not achieved within $T$ iterations.
After transformation, rows are renormalized to restore compositionality, 
inflating the relative abundances of the remaining taxa and altering the 
global compositional structure. This perturbation, therefore, jointly 
tests robustness to increased zero-inflation and to compositional 
redistribution.

\paragraph{{\densification} (\textit{Algorithm 3}.)}
It directly targets the zero-inflation structure of 
microbiome data, but in the opposite 
direction to {\sparsity}: rather than introducing new zeros, it 
fills existing ones.
Given a target sparsity level $\rho^\star$, we compute 
$k = \rho - \rho^\star$, where $\rho$ is the current number of zeros, as the number of zeros to fill, capped at 
$|\mathcal{Z}|$, the number of zero entries available in $\mathcal{F}_U$. 
We randomly select $k$ positions from $\mathcal{Z}$ and replace each 
zero entry $(i,j)$ with a value drawn uniformly from $\mathcal{V}_j$, 
the set of observed non-zero values of feature $j$ across all samples. 
This ensures that filled values are not synthetic noise but plausible 
abundance levels already observed in the dataset for that taxon. If 
feature $j$ has no observed non-zero values, a fallback feature is 
selected uniformly at random from $\mathcal{F}_U$. 
After filling, rows are renormalized to restore compositionality. 
As with the previous algorithms, this renormalization alters the global compositional structure beyond the mere injection of non-zero values, jointly analyzing robustness to reduced zero-inflation and to compositional redistribution.

\subsection{Evaluation pipeline}
\label{sec:evaluation}
\subsubsection{Query distribution shift}
We evaluate TFMs under a setting we name \textit{query distribution shift}: the support set $\mathcal{S}$ is always drawn from the original unperturbed data while the query set $\mathcal{Q}$ is replaced by its perturbed counterpart $\mathcal{Q}^\prime$ obtained via the algorithms we define in~\Cref{sec:perturbations}. The distribution shift is therefore defined with respect to the support, and its magnitude is controlled by the parameter of the perturbations (e.g., $k$, $\rho^\star$, $\tau$).

This setting differs from two related notions: adversarial attacks~\cite{goodfellow2014explaining} and out-of-distribution (OOD) generalization~\cite{hendrycks2016baseline}. In adversarial settings, perturbations are typically optimized to maximize model error under norm constraints and need not correspond to biologically meaningful transformations of the data. In contrast, OOD generalization studies scenarios where the test distribution differs from the distribution used to train the model. In our ICL setting, however, there is no task-specific training phase: the model receives $\mathcal{S}$ as context at inference time and must generalize to $\mathcal{Q}^\prime$ in a single forward pass. As a result, the distribution shift arises between the support and query sets rather than between training and test distributions.

\subsubsection{Why perturbing only the query}
Each perturbation family is designed to approximate a distinct source 
of support-query mismatch that may arise in clinical microbiome datasets. 
{\remove} captures scenarios in which features that are observed 
in the reference cohort are absent in the new samples, for instance, due to 
differences in sequencing depth, measurement procedures, or preprocessing 
pipelines. {\sparsity} models an increase in zero-inflation at 
query time, reflecting settings where query samples contain fewer 
observable taxa or where the measurement process produces sparser 
observations. {\densification} represents the opposite situation, where 
features absent in the support set become observable in the query, which 
may occur when models are applied across cohorts with higher measurement 
sensitivity or processed with different protocols.
In all three cases, the informative features $\mathcal{F}_I$ are 
preserved, ensuring that the discriminative signal underlying the task 
remains available, and that performance differences can be attributed 
to the distributional mismatch rather than to the removal of task-relevant 
information.
\section{Experiments}
The proposed framework comprises $6\ \text{datasets} \times 6\ \text{models} \times 3\ \text{perturbation families} \times 15\ \text{perturbation levels} \times 5\ \text{CV folds} = 8{,}100$ evaluations in total.
\subsection{Experimental setting}
\label{sec:exp_setting}
\subsubsection{Datasets}
We consider publicly curated metagenomic datasets from ExperimentHub~\cite{pasolli2017accessible}, which includes six cohorts with data at six taxonomic levels. We focused on the species-level taxa (microbial species relative abundance) across five cohorts: Cirrhosis (discovery: 178 samples, 83/95 control/case, 520 features; validation: 54 samples, 31/23, 394 features), Inflammatory Bowel Disease (IBD; 110 samples, 85/25, 443 features), Obesity (253 samples, 89/164, 465 features), Type 2 Diabetes in Chinese patients (T2D; 344 samples, 174/170, 572 features), and Type 2 Diabetes in European women (WT2D; 96 samples, 43/53, 381 features). 

For each cohort, the task is binary classification. No preprocessing was applied to any dataset.
\subsubsection{Perturbation pipeline parameter setting}
We described the selection process for informative and uninformative features in~\Cref{sec:perturbations}. Specifically, we perform 5-fold stratified cross-validation and set the $\delta$ threshold to $a_0 - a_j < \delta$ = 0.03, meaning the AUROC degrades by at most 3 percentage points. This threshold represents a reasoned balance: a $5\%$ degradation would suggest that a substantial biological signal has been removed, while a more stringent threshold would protect very few features, providing insufficient sensitivity to perturbations. Additional discussion is provided in Section C.1 (of supplementary materials).
\subsubsection{Perturbation algorithms parameter setting}
We described the perturbation algorithms in~\Cref{sec:perturbations}. In particular because of the heterogeneity of microbiome datasets, we set the perturbation parameters in a data-dependent manner, anchored to the original sparsity $\rho$ and the number of taxa $d$ of each dataset. 
\begin{description}
    \item[\textbf{Algorithm 1:}] the number of removed features $k$ ranges over a grid of 10 evenly spaced values between 1 to $d/2$. 
    \item[\textbf{Algorithm 2:}] the target sparsity is varied over a grid of 5 evenly spaced values between the original sparsity $\rho$ and $0.99$ (we increase the number of zeros). The binary search runs for at most $T = 50$ iterations with $\gamma_{\min} = 0$, $\gamma_{\max} = 10$, and 
detection threshold $\tau = 10^{-6}$. The upper bound $\gamma_{\max} = 10$ ensures sufficient shrinkage coverage: for any relative abundance $x < 10^{-6/11} \approx 0.285$, we have $x^{11} < \tau = 10^{-6}$, so the entry is set to zero. Since microbiome abundance data are compositional and our datasets 
contain hundreds of taxa, the vast majority of non-zero values fall below this threshold in practice. Moreover, when the number of available non-zero entries in $\mathcal{F}_U$ is insufficient to reach the target, i.e. $\phi_U < \rho^\star - \rho$, the algorithm clips $k$ to $\phi_U$, and the resulting sparsity will be lower than $\rho^\star$. In our experiments, this situation does not arise, as all datasets 
contain sufficiently many non-zero entries in $\mathcal{F}_U$ relative to the perturbation targets.
Finally, the algorithm terminates early once $|\rho_U' - k| \leq 1$, i.e., when the number of newly introduced zeros differs from the target by at most one entry. This condition is met well before $T$ iterations in all our experiments, confirming that $T = 50$ acts as a reasonable upper bound on the number of iterations. 
    \item[\textbf{Algorithm 3:}] the target sparsity is varied over a grid of 5 evenly spaced values between $0.01$ and $\rho$ (we reduce the number of zeros).
\end{description}

\subsubsection{TFMs}
\label{subsec:versions}
We evaluate four TFMs together with two classical machine-learning baselines: Random Forest (RF)~\cite{breiman2001random} and XGBoost~\cite{chen2016xgboost}, both configured with 100 estimators, as detailed in~\Cref{sec:evaluation}. All methods are assessed using 5-fold stratified cross-validation to ensure that class proportions are preserved across folds.
As for the TFMs, we consider the following models introduced in~\Cref{sec:related}. TabPFN~\cite{grinsztajn2025tabpfn,hollmann2022tabpfn,hollmann2025accurate} (v\texttt{6.3.2}) is run with its default settings, including the 3-model ensemble and no gradient computation. TabICL~\cite{qu2025tabicl,qu2026tabiclv2} (v\texttt{2.0.1}) is also used with default parameters, with 32 estimators for ensemble predictions. TabDPT~\cite{ma2024tabdpt} (v\texttt{1.1.12}) is included as well with 8 estimators as default; this model supports up to 100 features, applying PCA-based reduction when needed. 
ContextTab~\cite{spinaci2025contexttab} (v\texttt{1.0.1}) is used with a \texttt{context\_size} of 2048 and a \texttt{bagging\_factor} of 1.


\section{Results}
\label{sec:results}

\begin{figure}
  \centering

  \begin{subfigure}{\linewidth}
    \includegraphics[width=\linewidth]{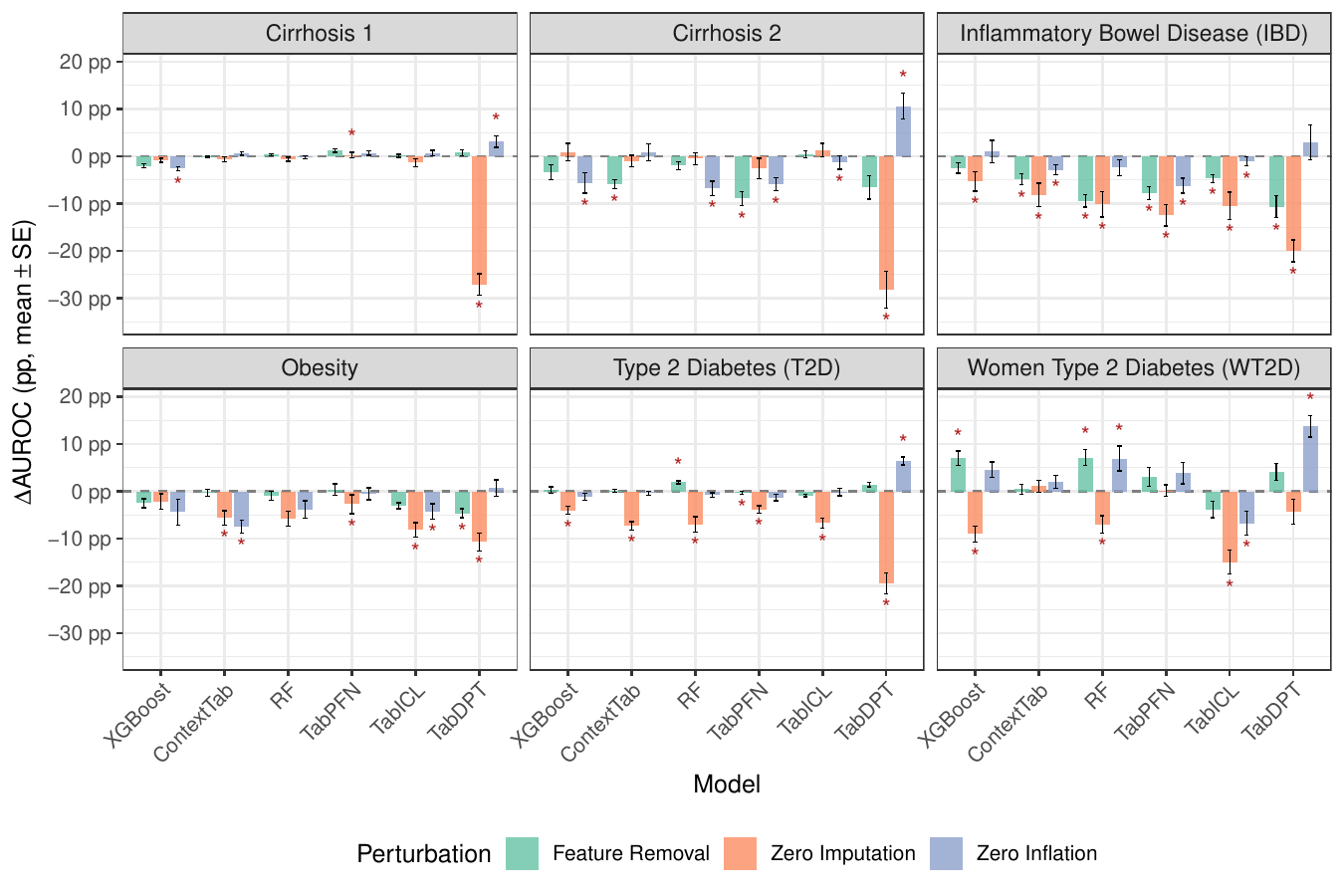}
    \subcaption{\textbf{AUROC degradation under perturbation} -- Mean change in AUROC (percentage points) relative to baseline for each model across datasets, grouped by perturbation. Error bars show ±1 standard error across cross-validation folds and perturbation levels. Asterisks denote model–dataset–perturbation combinations where at least one perturbation level had a statistically significant AUROC change \textit{(DeLong test, p < 0.05)}.}
    \label{fig:auroc_degradation}
  \end{subfigure}

  \vspace{0.3cm}

  \begin{subfigure}{\linewidth}
    \includegraphics[width=\linewidth]{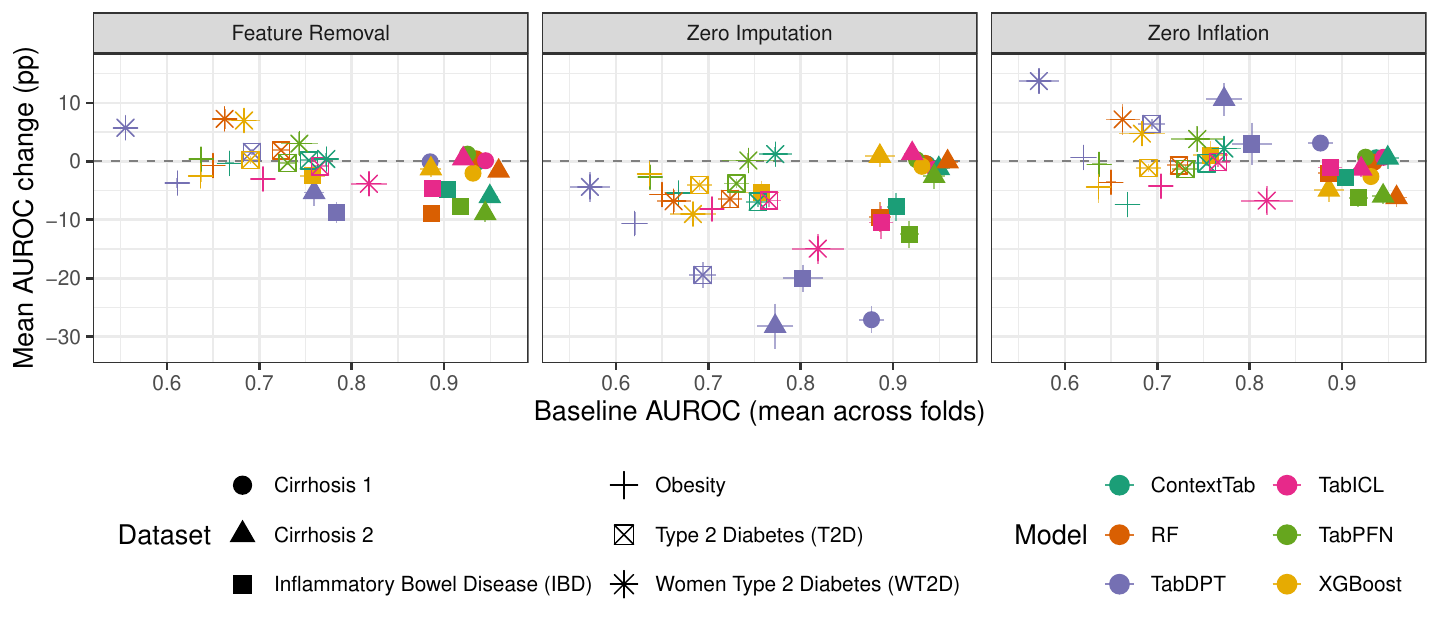}
    \subcaption{\textbf{Robustness–accuracy tradeoff} -- Relationship between baseline AUROC and mean AUROC change under perturbation for each model–dataset combination. Each point represents one model on one dataset. Error bars show ±1 standard error on both axes. Points near the dashed line \textit{(y = 0)}
   are robust, points in the upper-right quadrant combine high baseline accuracy with minimal degradation.}
    \label{fig:robustness_accuracy_tradeoff}
  \end{subfigure}

  \caption{\textbf{Model robustness under compositional perturbations.}
    (a)~AUROC degradation by model and perturbation type. (b)~Baseline accuracy versus robustness tradeoff per perturbation type.}
  \label{fig:main}
\end{figure}

We assess model robustness using the algorithms in~\Cref{sec:perturbations}, which simulate distribution shifts commonly faced when deploying microbiome classifiers across cohorts, sequencing platforms, or clinical sites. \Cref{fig:auroc_degradation} summarizes the mean AUROC degradation across all datasets and perturbations, and~\Cref{fig:robustness_accuracy_tradeoff} characterizes the tradeoff between baseline discriminative performance and robustness. Baseline AUROC values are reported in Table 2 (of supplementary materials). Throughout this section, all significance tests use the DeLong test~\cite{delong1988comparing} at $\alpha = 0.05$, and we denote by $n_{\text{sig}}$ the number of model--dataset--perturbation-level combinations reaching significance.
Additional results in Section C.3 (of supplementary materials).

\subsubsection{{\remove}}
Removing the most abundant (uninformative) features has a generally mild effect on AUROC (\Cref{fig:auroc_degradation}, green bars), which is expected given the high taxonomic redundancy of microbiome profiles. The most significant results fall under IBD, where all models degrade monotonically with the number of removed features~$k$. This suggests that the discriminative signal in IBD is spread across many taxa rather than carried by a few dominant species. \textit{TabDPT} shows the steepest drop on IBD ($-13$~pp), followed by \textit{RF} and \textit{TabPFN}. Only \textit{ContextTab} and \textit{TabPFN} on Cirrhosis~2 show sensitivity to feature removal.

\begin{figure}[h!]
    \includegraphics[width=\linewidth]{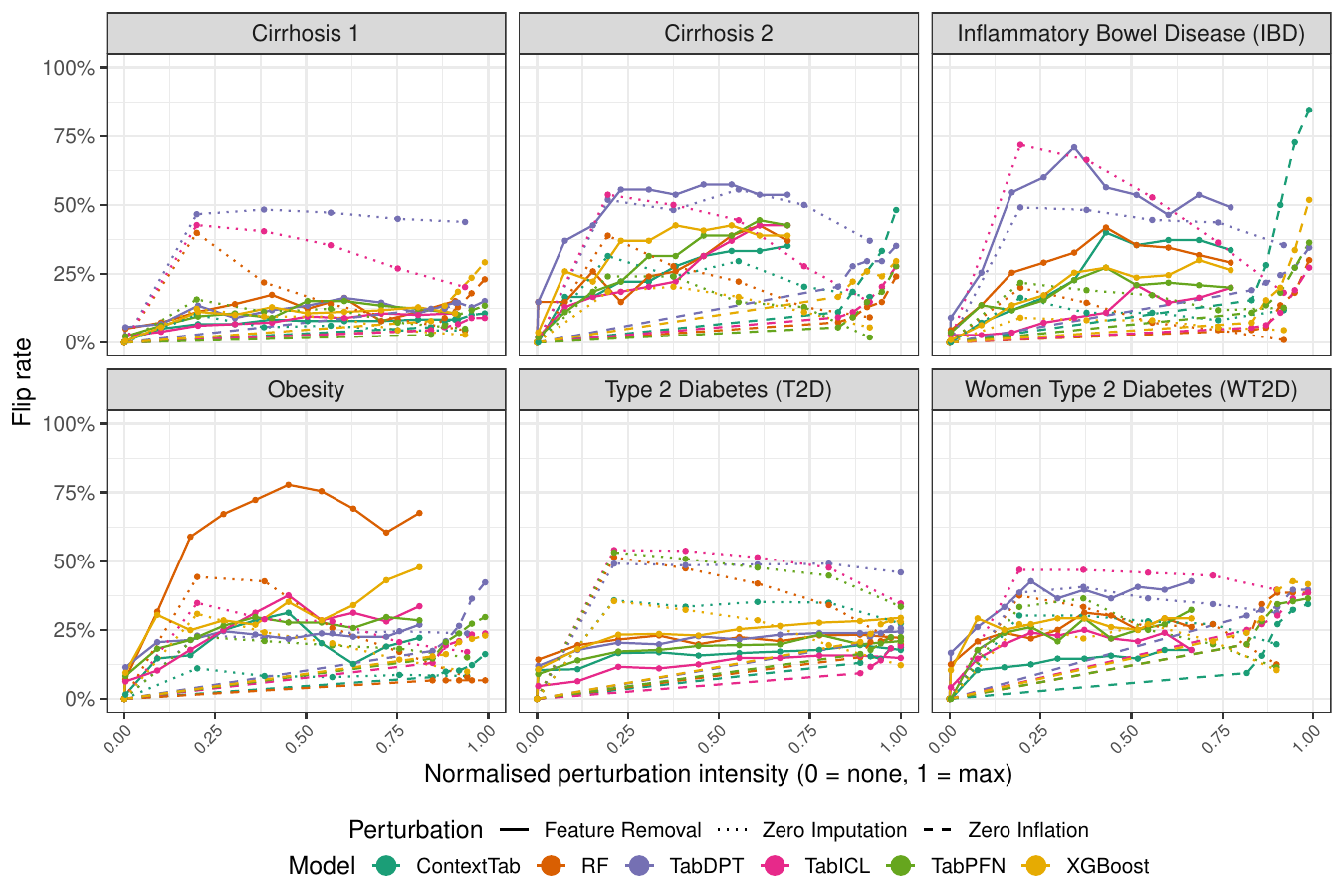}
    \caption{\textbf{Prediction flip rate under perturbation} -- Fraction of test samples whose predicted class changes relative to baseline as a function of normalised perturbation intensity \textit{(0 = unperturbed, 1 = maximum perturbation)}. A higher flip rate indicates greater decision instability — the
  model would assign a different classification to the same patient depending on data quality.}
    \label{fig:prediction_flip_rate}
\end{figure}

Interestingly, removal can also \emph{help}. Both tree-based models improve on WT2D, with mean gains of $+7.4$~pp (\textit{RF}) and $+5.9$~pp (\textit{XGBoost}) across perturbation levels, reaching up to $+11.0$~pp and $+11.6$~pp, respectively, at individual levels ($p < 0.05$). This is consistent with overfitting to noisy high-abundance features that the perturbation eliminates. For \textit{XGBoost}, this is the only significant result under feature removal across all datasets, suggesting that split-based feature selection already provides implicit regularisation. \textit{TabDPT} shows a similar trend on datasets with more features than its fixed input capacity ($d_{\text{feat}}$), where removing features reduces the need to compress the input and may preserve more signal, though the gains do not reach significance. These patterns are also visible at the individual prediction level. The flip rate (\Cref{fig:prediction_flip_rate}, solid lines) rises steeply for IBD, reaching ${\sim}55\%$ of predictions changed at maximum perturbation, whereas Cirrhosis~1 stays below 15\% for all models, confirming the stability of its discriminative signal. This contrast may also be partly amplified by class balance: IBD's pronounced imbalance (85/25) implies as few as ${\sim}5$ minority-class samples per test fold under 5-fold stratified CV, whereas Cirrhosis~1 is nearly balanced (83/95), yielding comparatively more stable per-fold estimates (Table 2 of supplementary materials). Overall, with respect to {\remove}, the results suggest that robustness is primarily determined by the structure of the discriminative signal in each dataset rather than by model architecture.  

\subsubsection{{\densification}}
While feature removal affects models moderately, zero imputation is far more damaging as it fills structural zeros with realistic non-zero values. No model improves under any condition; all significant results are degradations. \Cref{fig:robustness_accuracy_tradeoff} (\textit{center panel}) shows the widest vertical spread of the three perturbation types, reflecting large differences in how models respond.

\textit{TabDPT} stands out with 23 significant degradations, losing up to $-31.6$~pp on Cirrhosis~1 and $-32.1$~pp on Cirrhosis~2. In the scatter plot, its points lie well below those of all other models, regardless of baseline AUROC. \textit{TabDPT} operates as an in-context learner: at inference time, it conditions its predictions on a set of training examples (the context) passed alongside the test sample. The transformer's attention mechanism implicitly retrieves the most relevant context examples to inform its output. Because \textit{TabDPT} was pre-trained on data with many structural zeros, this retrieval works best when test inputs are similarly sparse. Zero imputation fills those zeros with realistic non-zero values, making the test input look fundamentally different from the context examples the model relies on. The attention mechanism can no longer match the test sample to relevant training patterns, resulting in large performance drops.

\textit{TabICL} is the next most affected ($n_{\text{sig}} = 16$). Its column-wise Set Transformer estimates the distribution of each feature across samples and uses these per-feature summaries to build row-level representations. As more zeros are replaced with imputed values, the observed feature distributions shift further from the true ones, and \textit{TabICL}'s representations become increasingly unreliable, which explains why its degradation grows monotonically with the fraction of imputed values. \textit{ContextTab} ($n_{\text{sig}} = 12$) degrades mainly on IBD, Obesity, and T2D, while \textit{RF}, \textit{TabPFN}, and \textit{XGBoost} are affected more selectively.

From the dataset side, T2D is the most vulnerable cohort ($n_{\text{sig}} = 25$, all six models degrade significantly), followed by IBD ($n_{\text{sig}} = 17$). Cirrhosis~1, by contrast, yields only 6 significant degradations, all from \textit{TabDPT}, and sits in the upper-right quadrant of the scatter plot. Its high density of informative features likely buffers against the introduction of spurious non-zero values.

\Cref{fig:prediction_flip_rate} (dotted lines) confirms how severe this perturbation is at the sample level: \textit{TabDPT} exceeds 50\% on Cirrhosis~2, and even \textit{XGBoost}, the most stable model overall, reaches 15--25\% on IBD and T2D. This shows that zero imputation affects not only aggregate metrics but also individual patient-level decisions.

This perturbation targets a central ambiguity of metagenomic data: zeros may denote true biological absence or technical dropout driven by sequencing depth or protocol differences~\cite{kumar2024comprehensive}. Our results, however, indicate that the observed degradation can be primarily attributable to model-side weaknesses rather than to this biological ambiguity: RF and XGBoost remain substantially more robust under zero imputation (\Cref{fig:auroc_degradation}), while TabDPT, TabICL, and ContextTab lose up to 30+ AUROC points. This asymmetry points to an architecture-specific reliance on the distributional structure of $\mathcal{F}_U$ - for instance, attention-based retrieval or per-feature distribution estimation, as the primary driver of the observed fragility - though ruling out alternative explanations entirely would require additional analyses motivated by the limitations of Zero Imputation discussed in (\Cref{sec:future}).

\subsubsection{{\sparsity}}
This perturbation increases sparsity by replacing non-zero abundance values with zeros. \textit{TabDPT} is the only model to improve significantly ($n_{\text{sig}} = 11$): up to $+19.7$~pp on WT2D, $+13.6$~pp on Cirrhosis~2, and $+8.8$~pp on T2D, while all other models degrade moderately. In the scatter plot, \textit{TabDPT} points are the only ones consistently above the zero line, confirming that increasing sparsity moves the input towards the sparse patterns of its training input, improving retrieval quality.  

Conversely, the models that were relatively resilient to zero imputation now degrade. \textit{ContextTab} loses up to $-12.1$~pp on Obesity ($n_{\text{sig}} = 4$) and $-11.0$~pp on IBD, while \textit{TabICL} drops up to $-9.9$~pp on Obesity ($n_{\text{sig}} = 3$), $-11.2$~pp on Cirrhosis~2, and $-10.0$~pp on WT2D. Despite differing architectures, both models share a key property: they have learned realistic abundance distributions during pre-training. \textit{TabICL} does so through its column-wise Set Transformer, which builds feature-distribution-aware embeddings, \textit{ContextTab} relies on column-name embeddings that couple feature identity with abundance patterns. 
Zero inflation destroys precisely these learned distributions by replacing informative values with zeros. \textit{TabPFN}, despite sharing a dual-attention architecture with \textit{ContextTab}, is comparatively resilient -- trained exclusively on synthetic data, it carries no prior on real feature distributions. Tree-based models degrade modestly and only on specific datasets. Cirrhosis~1, as under feature removal, shows no significant degradation for any model.

Flip rates (\Cref{fig:prediction_flip_rate}, dashed lines) vary strongly across datasets. IBD reaches ${\sim}85\%$ for \textit{ContextTab} at maximum perturbation, while T2D and Obesity remain below 30\% for most models, likely reflecting differences in the fraction of biologically meaningful non-zero entries.
This perturbation, therefore, shows a genuine fragility: models are sensitive to changes in abundance values that should carry negligible discriminative weight.

\subsubsection{Cross-model shift concordance}
\begin{figure}[t!]
    \includegraphics[width=\linewidth]{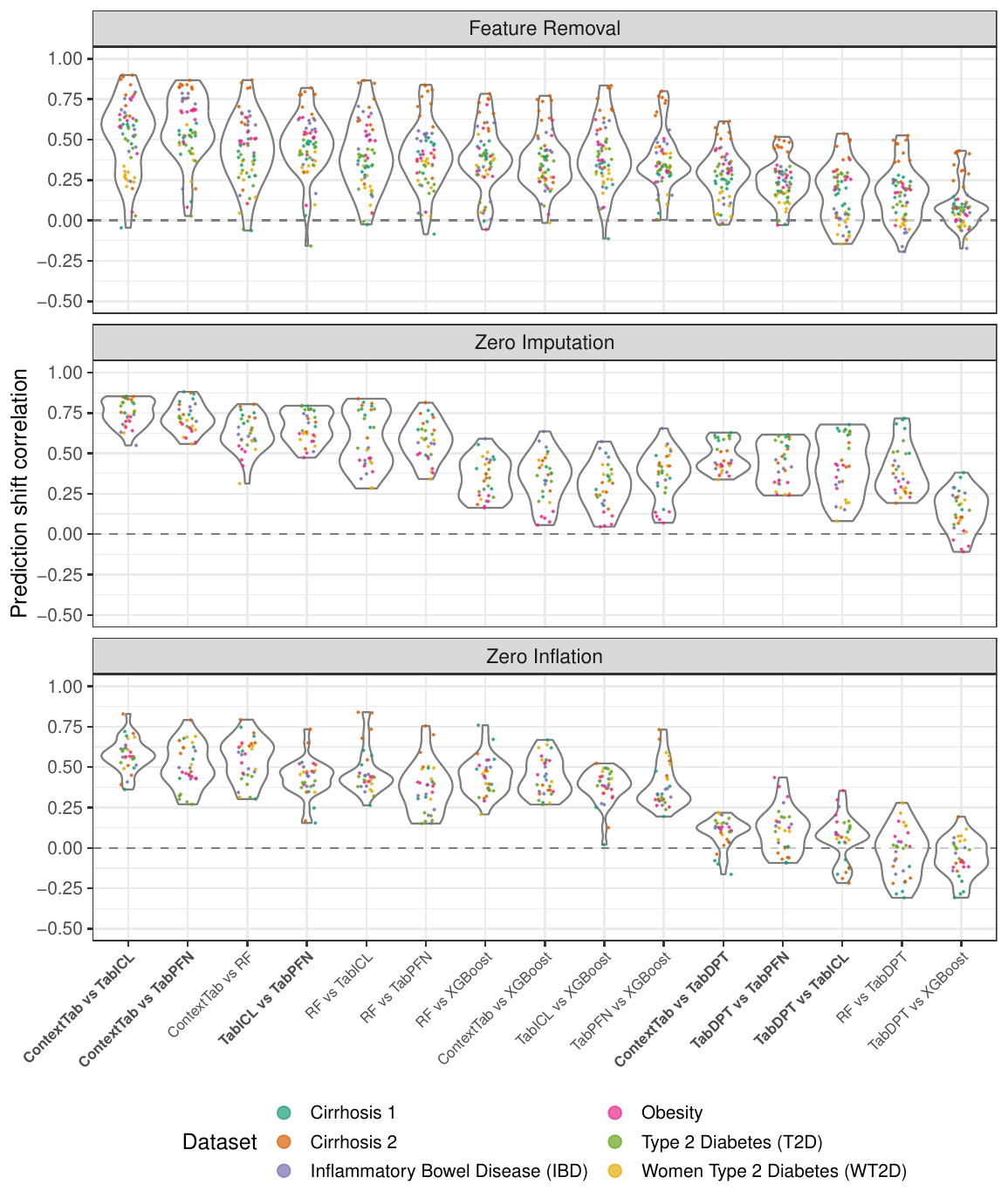}
    \caption{\textbf{Pairwise prediction shift concordance} -- Distribution of Spearman correlations between per-sample prediction shifts across all model pairs, faceted by perturbation type. Each point represents a dataset at a given perturbation level. High positive correlations indicate that both models are affected in the same direction on the same samples \textit{(shared vulnerabilities)}, low or negative correlations indicate complementary failure modes, favorable for ensembling.}
    \label{fig:shift_concordance}
\end{figure}
The results above do not reveal whether models fail on the same samples. To answer this question, we compute pairwise Spearman correlations of per-sample prediction shifts across all perturbation levels and datasets (\Cref{fig:shift_concordance}). The results indicate a clear stratification. ICL-based foundation model pairs (\textit{ContextTab}--\textit{TabICL}, \textit{ContextTab}--\textit{TabPFN}, \textit{TabICL}--\textit{TabPFN}) and the \textit{ContextTab}--\textit{RF} pair show the highest concordance (median $\rho \approx 0.6$--$0.85$) across all three perturbation types: when one model shifts its prediction on a sample, the other shifts in the same direction. These models share failure modes at the instance level.
All pairs involving \textit{TabDPT} cluster at the bottom, with median correlations near zero and sometimes negative under zero inflation. This means \textit{TabDPT} fails on different samples and often in the opposite direction compared to other models. Despite its fragility under zero imputation, this complementarity makes it a valuable candidate for ensemble strategies. Importantly, this concordance structure is stable across perturbations: pairs that are concordant under feature removal remain concordant under zero imputation and zero inflation. This suggests that the similarity of failure patterns reflects shared architectural priors, in particular the ICL paradigm, rather than properties of the perturbation. In practice, the ensembling potential of a model pair can therefore be assessed from a single perturbation type.

\section{Conclusions}
We proposed a framework to evaluate TFMs on microbiome data under realistic query distribution shifts. Using three biologically motivated perturbation strategies: {\remove}, {\sparsity} and {\densification}, applied across six gut microbiome datasets and six models, we characterized how structural properties unique to microbiome data challenge in-context learners in ways that standard benchmarks do not reveal.

We report three main findings. First, \textbf{no model is robust across all perturbation types}: every model degrades significantly under at least one condition, and the pattern of vulnerabilities differs across architectures. {\densification} is consistently the most harmful perturbation, causing widespread degradation across all models and datasets, whereas {\remove} has comparatively mild, dataset-dependent effects. Second, \textbf{protecting discriminative features is insufficient to guarantee stability}: even when informative taxa are fully preserved, corrupting the global compositional and sparsity structure of query samples degrades in-context learning, indicating that TFMs rely on global feature structure beyond the discriminative signal itself. Third, \textbf{there is a tradeoff between semantic richness and distributional robustness}. Models that have learned realistic abundance distributions during pre-training -- such as \textit{TabICL} and \textit{ContextTab} -- achieve stronger baseline performance but are disproportionately sensitive to zero inflation type shifts. Conversely, \textit{TabPFN} exhibits greater robustness to sparsity perturbations at the cost of richer representational capacity. \textit{TabDPT} occupies a complementary niche: severely vulnerable to zero imputation yet uniquely benefiting from increased sparsity, with failure modes largely uncorrelated with other models.

\textbf{From a practical standpoint, dataset properties matter as much as model choice}. Models on the Cirrhosis~1 dataset remain robust under all perturbations, unlike in T2D and IBD, suggesting that the density and redundancy of informative features contribute to robustness. When sparsity patterns or cohort composition may differ between training and deployment, as is common in multi-site microbiome studies, simpler models or architecturally diverse ensembles may offer better reliability than any single TFM.
Our findings also raise questions about interpretability and clinical relevance, as model-based feature importance may not reflect biological function. Taxa considered uninformative by our criterion may still be clinically meaningful, while model-relied features may capture cohort artefacts rather than true biological signal.






\section{Future Work}
\label{sec:future}

\noindent\textbf{Improving robustness.}
Our findings motivate several directions for future work. The most immediate 
is to move from characterizing fragility to mitigating it. Fine-tuning TFMs 
on metagenomic cohorts would align their priors with the compositional, 
zero-inflated structure that synthetic pretraining does not capture, while 
data augmentation exposing models to support-query sparsity mismatch during 
training, together with post-hoc calibration, offers complementary routes 
toward greater stability under distribution shift.

\noindent\textbf{Data representation.}
We evaluated models on raw relative abundances, leaving compositional
transforms such as the centered log-ratio (CLR) and its robust variants
unexplored~\cite{martino2019novel}. Whether such transforms help or hinder TFM representations
remains an open question: these models apply their own internal
normalization, which an external log-ratio transform might disrupt
rather than complement.

\noindent\textbf{Perturbation refinement.}
Zero Imputation fills structural zeros by sampling from each feature's
empirical non-zero distribution. However, technical false zeros (taxa
present but undetected owing to limited sequencing depth) are more
plausibly located just above the detection limit; restricting sampling to
the lower tail of each feature's distribution would therefore yield a more
biologically faithful densification and constitutes a natural refinement of Algorithm 3.

\noindent\textbf{Feature selection criterion.}
The partition into protected and perturbed features relies on ANOVA F-tests
and a Random Forest wrapper, which may not capture the non-linear
dependencies that foundation models exploit. A model-agnostic attribution
method such as SHAP, or controlled experiments with a known ground-truth
signal, would help clarify whether the measured degradation reflects genuine
robustness gaps or artifacts of the importance criterion. Such analyses would
also help bridge the gap between data-driven importance and biological
relevance, since taxa deemed uninformative by our criterion may nonetheless
be clinically meaningful.

\noindent\textbf{Benchmark extension and biological validation.}
The benchmark could also be extended with larger context sizes, alternative 
support-query compositions, and a broader model pool --- though the Pasolli 
cohorts, spanning four disease contexts and six datasets of varying size, 
class balance, and sparsity structure, already provide a representative 
testbed for the biological variability of interest. Targeted biological 
validation --- for instance, datasets with experimentally verified absent 
taxa --- would further strengthen the cross-model evidence reported in 
\Cref{sec:results} and help establish whether the sparsity patterns that TFMs 
rely on reflect genuine biological signal or cohort-specific artifacts.

\begin{credits}
\subsubsection{\ackname}This work was supported by a grant from the French Agence Nationale de la Recherche (ANR) for the DeepIntegrOmics project (number ANR-21-CE45-0030).
This work was granted access to the HPC resources of IDRIS (Institut du développement et des ressources en informatique scientifique) under the allocations 2023-AD011014580, 2024-AD011014580R1 and 2024-AD011015723R1 made by GENCI (Grand Équipement National de Calcul Intensif).
\end{credits}

\bibliographystyle{splncs04}
\bibliography{biblio}

\begin{thebibliography}{10}
\providecommand{\url}[1]{\texttt{#1}}
\providecommand{\urlprefix}{URL }
\providecommand{\doi}[1]{https://doi.org/#1}

\bibitem{breiman2001random}
Breiman, L.: Random forests. Machine learning  \textbf{45}(1),  5--32 (2001)

\bibitem{busato2023compositionality}
Busato, S., Gordon, M., Chaudhari, M., Jensen, I., Akyol, T., Andersen, S.,
  Williams, C.: Compositionality, sparsity, spurious heterogeneity, and other
  data-driven challenges for machine learning algorithms within plant
  microbiome studies. Current Opinion in Plant Biology  \textbf{71},  102326
  (2023)

\bibitem{chen2016xgboost}
Chen, T., Guestrin, C.: Xgboost: A scalable tree boosting system. In:
  Proceedings of the 22nd acm sigkdd international conference on knowledge
  discovery and data mining. pp. 785--794 (2016)

\bibitem{delong1988comparing}
DeLong, E.R., DeLong, D.M., Clarke-Pearson, D.L.: Comparing the areas under two
  or more correlated receiver operating characteristic curves: a nonparametric
  approach. Biometrics pp. 837--845 (1988)

\bibitem{forry2024variability}
Forry, S.P., Servetas, S.L., Kralj, J.G., Soh, K., Hadjithomas, M., Cano, R.,
  Carlin, M., Amorim, M.G.d., Auch, B., Bakker, M.G., et~al.: Variability and
  bias in microbiome metagenomic sequencing: an interlaboratory study comparing
  experimental protocols. Scientific reports  \textbf{14}(1), ~9785 (2024)

\bibitem{gardner2024large}
Gardner, J., Perdomo, J.C., Schmidt, L.: Large scale transfer learning for
  tabular data via language modeling. Advances in Neural Information Processing
  Systems  \textbf{37},  45155--45205 (2024)

\bibitem{goodfellow2014explaining}
Goodfellow, I.J., Shlens, J., Szegedy, C.: Explaining and harnessing
  adversarial examples. arXiv preprint arXiv:1412.6572  (2014)

\bibitem{grinsztajn2025tabpfn}
Grinsztajn, L., Fl{\"o}ge, K., Key, O., Birkel, F., Jund, P., Roof, B.,
  J{\"a}ger, B., Safaric, D., Alessi, S., Hayler, A., et~al.: Tabpfn-2.5:
  Advancing the state of the art in tabular foundation models. arXiv preprint
  arXiv:2511.08667  (2025)

\bibitem{hendrycks2016baseline}
Hendrycks, D., Gimpel, K.: A baseline for detecting misclassified and
  out-of-distribution examples in neural networks. arXiv preprint
  arXiv:1610.02136  (2016)

\bibitem{hollmann2022tabpfn}
Hollmann, N., M{\"u}ller, S., Eggensperger, K., Hutter, F.: Tabpfn: A
  transformer that solves small tabular classification problems in a second.
  arXiv preprint arXiv:2207.01848  (2022)

\bibitem{hollmann2025accurate}
Hollmann, N., M{\"u}ller, S., Purucker, L., Krishnakumar, A., K{\"o}rfer, M.,
  Hoo, S.B., Schirrmeister, R.T., Hutter, F.: Accurate predictions on small
  data with a tabular foundation model. Nature  \textbf{637}(8045),  319--326
  (2025)

\bibitem{kolberg2025tabpfn}
Kolberg, C., Eggensperger, K., Pfeifer, N.: Tabpfn-wide: Continued pre-training
  for extreme feature counts. arXiv preprint arXiv:2510.06162  (2025)

\bibitem{kumar2024comprehensive}
Kumar, B., Lorusso, E., Fosso, B., Pesole, G.: A comprehensive overview of
  microbiome data in the light of machine learning applications:
  categorization, accessibility, and future directions. Frontiers in
  microbiology  \textbf{15},  1343572 (2024)

\bibitem{ma2024tabdpt}
Ma, J., Thomas, V., Hosseinzadeh, R., Labach, A., Kamkari, H., Cresswell, J.C.,
  Golestan, K., Yu, G., Caterini, A.L., Volkovs, M.: Tabdpt: Scaling tabular
  foundation models on real data. arXiv preprint arXiv:2410.18164  (2024)

\bibitem{martino2019novel}
Martino, C., Morton, J.T., Marotz, C.A., Thompson, L.R., Tripathi, A., Knight,
  R., Zengler, K.: A novel sparse compositional technique reveals microbial
  perturbations. MSystems  \textbf{4}(1),  10--1128 (2019)

\bibitem{muller2021transformers}
M{\"u}ller, S., Hollmann, N., Arango, S.P., Grabocka, J., Hutter, F.:
  Transformers can do bayesian inference. arXiv preprint arXiv:2112.10510
  (2021)

\bibitem{pasolli2017accessible}
Pasolli, E., Schiffer, L., Manghi, P., Renson, A., Obenchain, V., Truong, D.T.,
  Beghini, F., Malik, F., Ramos, M., Dowd, J.B., et~al.: Accessible, curated
  metagenomic data through experimenthub. Nature methods  \textbf{14}(11),
  1023--1024 (2017)

\bibitem{perciballi2024adapting}
Perciballi, G., Granese, F., Fall, A., Zehraoui, F., Prifti, E., Zucker, J.D.:
  Adapting tabpfn for zero-inflated metagenomic data. In: NeurIPS 2024 Third
  Table Representation Learning Workshop (2024)

\bibitem{qu2025tabicl}
Qu, J., Holzm{\~A}{\v{z}}ller, D., Varoquaux, G., Morvan, M.L.: Tabicl: A
  tabular foundation model for in-context learning on large data. arXiv
  preprint arXiv:2502.05564  (2025)

\bibitem{qu2026tabiclv2}
Qu, J., Holzm{\"u}ller, D., Varoquaux, G., Morvan, M.L.: Tabiclv2: A better,
  faster, scalable, and open tabular foundation model. arXiv preprint
  arXiv:2602.11139  (2026)

\bibitem{shreiner2015gut}
Shreiner, A.B., Kao, J.Y., Young, V.B.: The gut microbiome in health and in
  disease. Current opinion in gastroenterology  \textbf{31}(1),  69--75 (2015)

\bibitem{spinaci2025contexttab}
Spinaci, M., Polewczyk, M., Schambach, M., Thelin, S.: Contexttab: A
  semantics-aware tabular in-context learner. arXiv preprint arXiv:2506.10707
  (2025)

\bibitem{vanschoren2014openml}
Vanschoren, J., Van~Rijn, J.N., Bischl, B., Torgo, L.: Openml: networked
  science in machine learning. ACM SIGKDD Explorations Newsletter
  \textbf{15}(2),  49--60 (2014)

\bibitem{zeng2025tabflex}
Zeng, Y., Dinh, T., Kang, W., Mueller, A.C.: Tabflex: Scaling tabular learning
  to millions with linear attention. arXiv preprint arXiv:2506.05584  (2025)

\end{thebibliography}

\appendix
\section{Supplementary Material to Section 3}
\subsection{Perturbed data example}
\label{app:perturbed}
\begin{figure}[h!]
    \setcounter{figure}{4}
    \centering
    \begin{subfigure}{0.8\linewidth}
        \includegraphics[width=\linewidth]{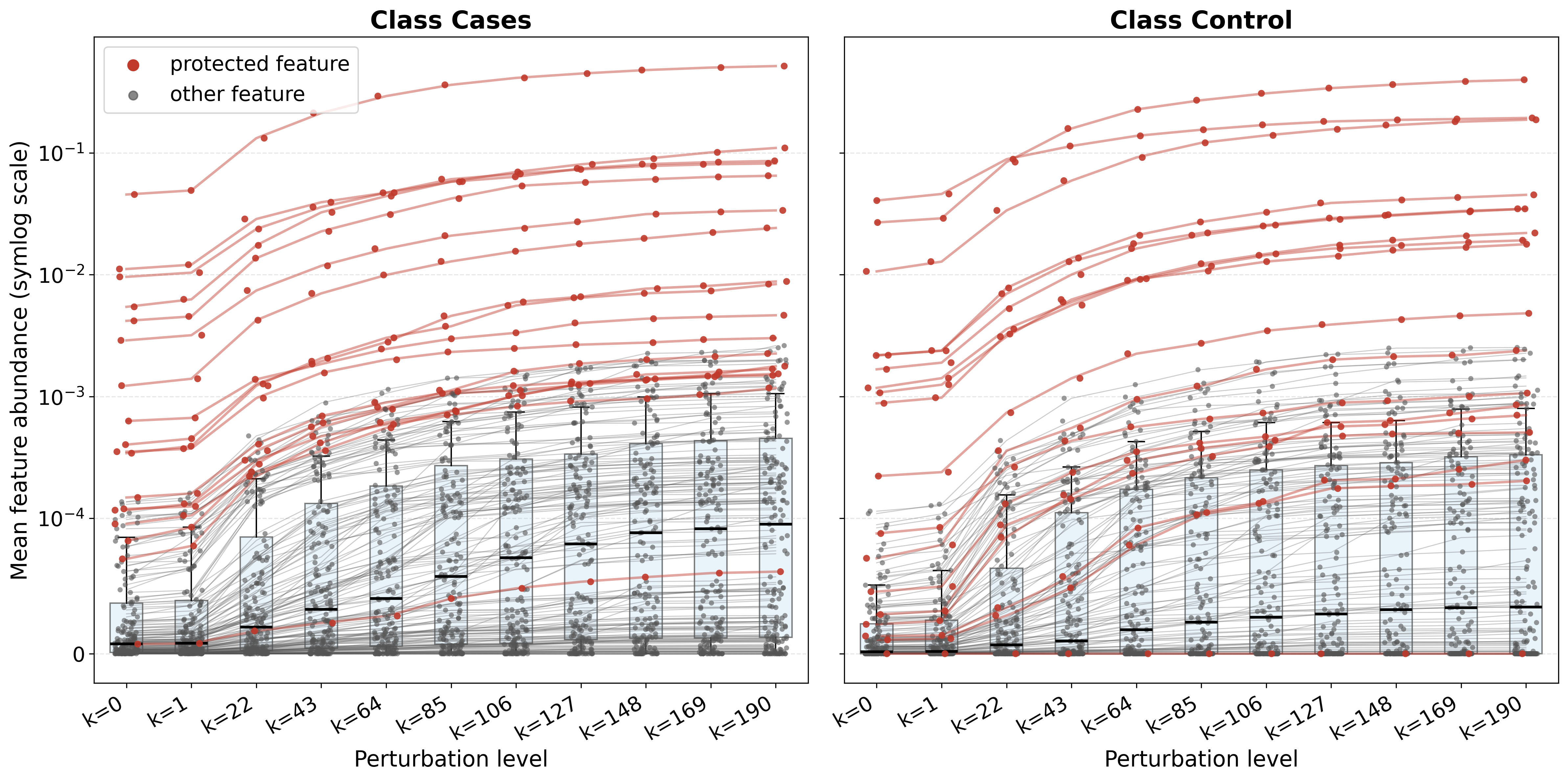}
        \caption{{\remove}}
        \label{fig:traj_remove}
    \end{subfigure}
    \hfill
    \begin{subfigure}{0.8\linewidth}
        \includegraphics[width=\columnwidth]{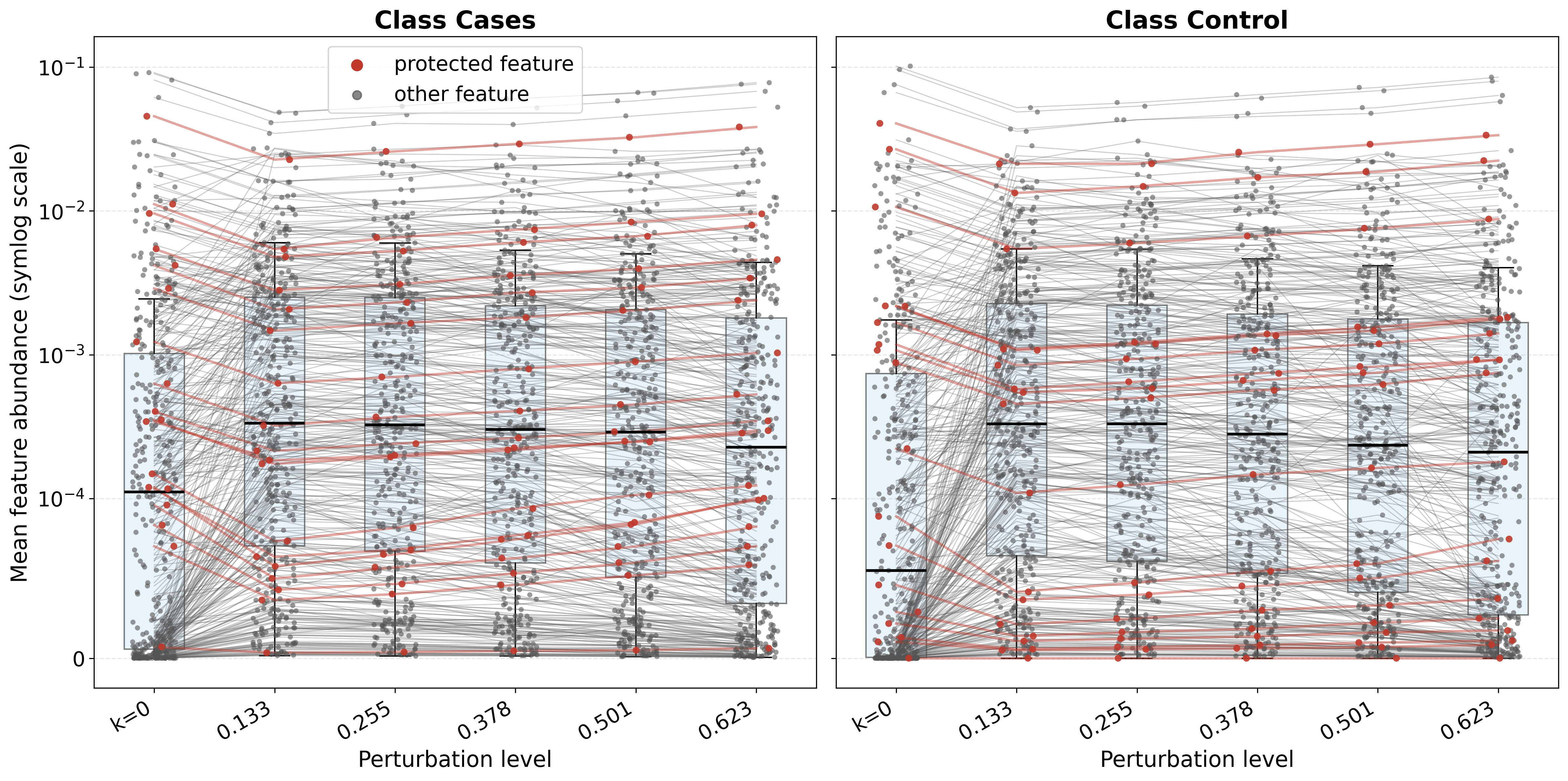}
        \caption{{\densification}}
        \label{fig:traj_sparsity}
    \end{subfigure}
    \hfill
    \begin{subfigure}{0.8\linewidth}
        \includegraphics[width=\columnwidth]{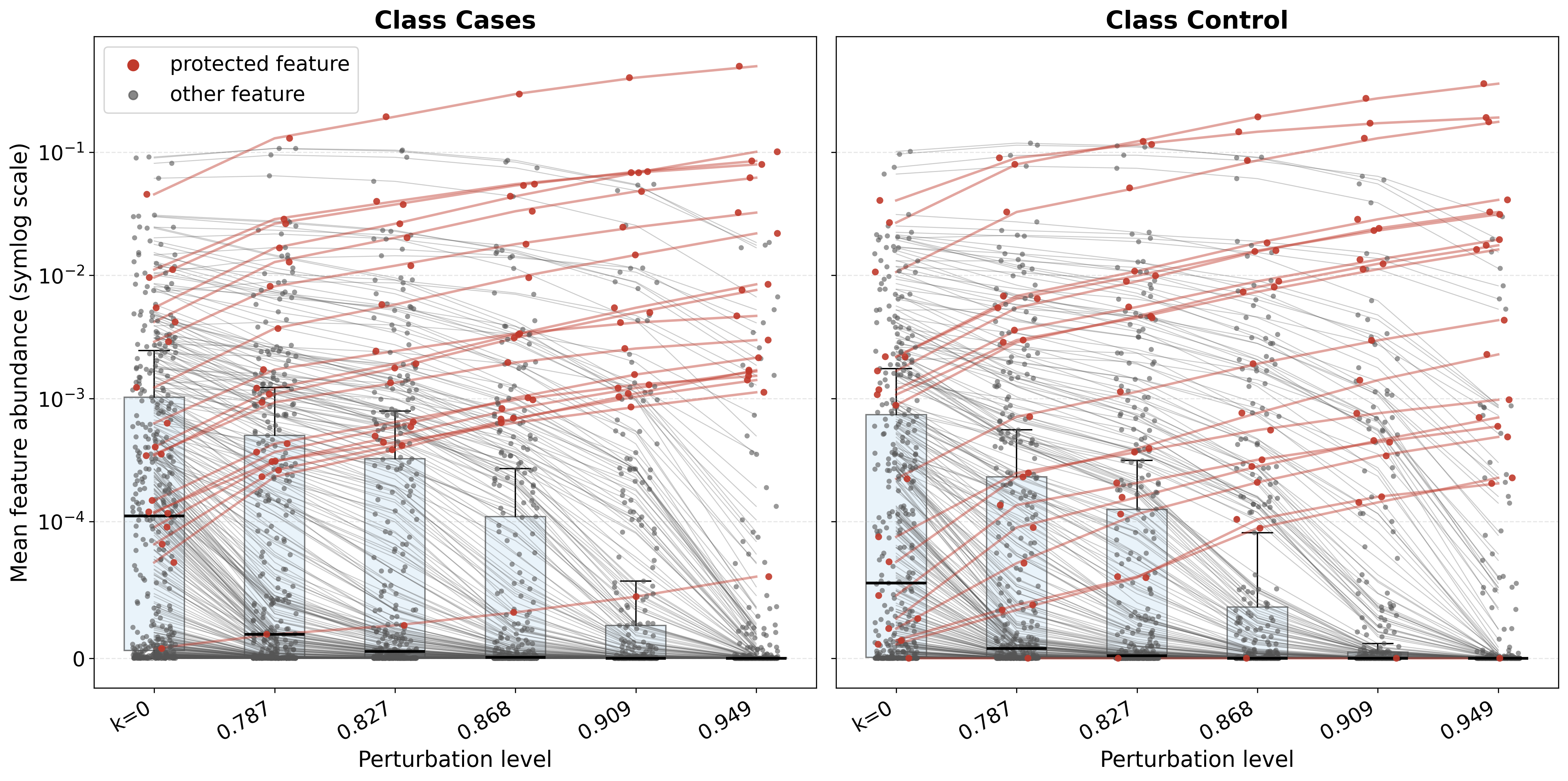}
        \caption{{\sparsity}}
        \label{fig:traj_sparsity}
    \end{subfigure}

    \caption{\textbf{Mean abundance distributions of individual features across increasing perturbation levels, stratified by class (Controls and Cases)} -- Red lines indicate ANOVA-selected protected features, which are preserved during perturbation and show consistent abundance enrichment relative to unprotected features (grey). Boxplots summarize the distribution of all features at each perturbation level; their parallel trajectories indicate that neither perturbation disrupts the relative class-level structure.}
    \label{fig:trajectories_both}
\end{figure}
\subsection{Perturbation algorithm pseudocodes}
\label{app:algorithms} 
\begin{algorithm}[htbp]
\caption{\remove}
\label{alg:remove_features}
\begin{algorithmic}[1]
\Require Taxonomic abundance matrix $\mathbf{X} \in \mathbb{R}^{n \times d}$, 
set of uninformative features $\mathcal{F}_{U}$, 
number of features to remove $k$.
\Ensure  Taxonomic abundance matrix $\mathbf{X}^\prime \in \mathbb{R}^{n \times (d-k)}$.
    \State $\mu_j \gets \frac{1}{n}\sum_{i=1}^{n} x_{ij}$ \quad $\forall\, j \in \mathcal{F}_U$
    \Comment{Compute mean abundance over $\mathcal{F}_U$}
    \State Sort features $j \in \mathcal{F}_U$ in descending order by $\mu_j$
    \State $\mathcal{F} \gets \{\texttt{sorted}(\mathcal{F}_U)\}_{1:k}$
\Comment{Select the $k$ highest-abundance uninformative features}
    \State $\mathcal{K} \gets \{1,\dots,d\} \setminus \mathcal{F}$
    \Comment{Retain all columns not selected for removal}
    \For{$i = 1$ to $n$}
        \State $s_i \gets \sum_{j \in \mathcal{K}} x_{ij}$
        \If{$s_i > 0$}
            \State $x'_{ij} \gets x_{ij} / s_i$ \quad $\forall\, j \in \mathcal{K}$
            \Comment{Re-normalize to restore compositionality}
        \Else
            \State $x'_{ij} \gets 0$ \quad $\forall\, j \in \mathcal{K}$
        \EndIf
    \EndFor
\State \Return $\mathbf{X}'$
\end{algorithmic}
\end{algorithm}

\begin{algorithm}[htbp]
\caption{\sparsity}
\label{alg:sparsify}
\begin{algorithmic}[1]
\Require Taxonomic abundance matrix $\mathbf{X} \in \mathbb{R}^{n \times d}$, 
         set of uninformative features $\mathcal{F}_{U}$, 
         target sparsity $\rho^\star_{\%} \in (0,1)$, 
         threshold $\tau = 10^{-6}$, 
         number of iterations $T$,
         search bounds $\gamma_{\min} \geq 0$, $\gamma_{\max} > \gamma_{\min}$.
\Statex \textbf{Convention:} Indices $i \in \{1,\dots,n\}$ denote samples, 
        $j \in \{1,\dots,d\}$ denote features.
\Ensure Taxonomic abundance matrix $\mathbf{X}^\prime \in \mathbb{R}^{n \times d}$ 
        with sparsity less than $\rho^\star_{\%}$.
\State $\rho \gets \sum_{i,j} \mathds{1}[x_{ij} = 0]$
\Comment{Current number of zeros in $\mathbf{X}$}
\State $\rho^\star \gets \lfloor \rho^\star_{\%} \cdot n \cdot d \rfloor$
\Comment{Target number of zeros}
\State $\phi_U \gets \sum_{i,j} \mathds{1}[x_{ij} > 0] \cdot \mathds{1}[j \in \mathcal{F}_U]$
\Comment{Non-zero entries available for sparsification}
\State $k \gets \min\{\rho^\star - \rho,\, \phi_U\}$
\Comment{Number of zeros to introduce}
\If{$k < 0$}
    \State \textbf{raise} \texttt{ValueError}: target sparsity $\rho^\star_{\%}$ 
    is already exceeded by the current sparsity
\EndIf
\If{$k = 0$}
    \State \Return $\mathbf{X}$ \Comment{No sparsification needed}
\EndIf
\State $\rho_U \gets \sum_{i,j} \mathds{1}[x_{ij} = 0] \cdot \mathds{1}[j \in \mathcal{F}_U]$
\Comment{Current zeros in $\mathcal{F}_U$}
\State $\mathbf{X}^* \gets \mathbf{X}$,\; $d^* \gets \infty$
\For{$t = 1$ to $T$}
    \State $\gamma \gets (\gamma_{\min} + \gamma_{\max}) / 2$
    \State $\tilde{x}_{ij} \gets x_{ij}$ \quad $\forall\, j \notin \mathcal{F}_U$
    \Comment{Informative features are left unchanged}
    \State $\tilde{x}_{ij} \gets (x_{ij})^{1+\gamma} \cdot \mathds{1}[(x_{ij})^{1+\gamma} \geq \tau]$ 
    \quad $\forall\, j \in \mathcal{F}_U$
    \Comment{Power transformation with threshold}
    \State $\rho_U^\prime \gets \sum_{i,j} \mathds{1}[\tilde{x}_{ij} = 0] \cdot 
    \mathds{1}[j \in \mathcal{F}_U] - \rho_U$
    \Comment{Newly introduced zeros}
    \If{$|\rho_U^\prime - k| < d^*$}
        \State $d^* \gets |\rho_U^\prime - k|$;\; 
        $x^*_{ij} \gets \tilde{x}_{ij}\ \forall\, i,j$
        \Comment{Save best iterate}
    \EndIf
    \If{$|\rho_U^\prime - k| \leq 1$} \textbf{break} \EndIf
    \If{$\rho_U^\prime < k$} $\gamma_{\min} \gets \gamma$ 
    \Else\ $\gamma_{\max} \gets \gamma$ \EndIf
\EndFor
\For{$i = 1$ to $n$}
\Comment{Re-normalize rows to restore compositionality}
    \State $s_i \gets \sum_{j=1}^{d} x^*_{ij}$
    \If{$s_i > 0$} 
        \State $x^\prime_{ij} \gets x^*_{ij} / s_i$ \quad $\forall\, j$
    \Else
        \State $x^\prime_{ij} \gets 0$ \quad $\forall\, j$
    \EndIf
\EndFor
\State \Return $\mathbf{X}^\prime$
\end{algorithmic}
\end{algorithm}

\begin{algorithm}[htbp]
\caption{\densification}
\label{alg:densification}
\begin{algorithmic}[1]
\Require Taxonomic abundance matrix $\mathbf{X} \in \mathbb{R}^{n \times d}$,
         set of uninformative features $\mathcal{F}_{U}$,
         target sparsity $\rho^\star_{\%} \in (0,1)$
         .
\Statex \textbf{Convention:} Indices $i \in \{1,\dots,n\}$ denote samples, 
        $j \in \{1,\dots,d\}$ denote features
\Ensure Taxonomic abundance matrix $\mathbf{X}^\prime \in \mathbb{R}^{n \times d}$ 
        with sparsity less than $\rho^\star_{\%}$
\State $\rho \gets \sum_{i,j} \mathds{1}[x_{ij} = 0]$
\Comment{Current number of zeros in $\mathbf{X}$}
\State $\rho^\star \gets \lfloor \rho^\star_{\%} \cdot n \cdot d \rfloor$
\Comment{Target number of zeros}
\State $k \gets \rho - \rho^\star$
\Comment{Number of zeros to fill}
\If{$k < 0$}
    \State \textbf{raise} \texttt{ValueError}: target sparsity $\rho^\star_{\%}$ 
    is already below the current sparsity
\EndIf
\If{$k = 0$}
    \State \Return $\mathbf{X}$ \Comment{No densification needed}
\EndIf
\State $\mathcal{Z} \gets \{(i,j) : x_{ij} = 0,\; j \in \mathcal{F}_U\}$
\Comment{Zero entries available for densification}
\State $k \gets \min\{k,\, |\mathcal{Z}|\}$
\Comment{Cap at available zero entries in $\mathcal{F}_U$}
\For{$j \in \mathcal{F}_U$}
\Comment{Collect observed non-zero values per feature}
    \State $\mathcal{V}_j \gets \{x_{ij} : x_{ij} > 0\}$
    \Comment{Non-zero values of feature $j$}
\EndFor
\If{$\forall\, j \in \mathcal{F}_U:\, |\mathcal{V}_j| = 0$}
    \State \Return $\mathbf{X}$ \Comment{No non-zero values available to sample from}
\EndIf
\State $\mathcal{S} \gets$ sample $k$ positions from $\mathcal{Z}$ uniformly at random
\State $\mathbf{X}^\star \gets \mathbf{X}$
\For{$\ell = 1$ to $k$}
    \State $(i,j) \gets \mathcal{S}_\ell$
    \If{$|\mathcal{V}_j| > 0$}
        \State $x^\star_{ij} \gets$ sample uniformly at random from $\mathcal{V}_j$
        \Comment{Draw from observed non-zero values of feature $j$}
    \Else
        \State $j^\star \gets$ sample uniformly from 
        $\{j^{\star\star} \in \mathcal{F}_U : |\mathcal{V}_{j^{\star\star}}| > 0\}$
        \State $x^\star_{ij} \gets$ sample uniformly at random from $\mathcal{V}_{j^\star}$
        \Comment{Fallback: draw from observed non-zero values of feature $j^\star$}
    \EndIf
\EndFor
\For{$i = 1$ to $n$}
\Comment{Re-normalize rows to restore compositionality}
    \State $s_i \gets \sum_{j=1}^{d} x^\star_{ij}$
    \If{$s_i > 0$}
        \State $x^\prime_{ij} \gets x^\star_{ij} / s_i$ \quad $\forall\, j$
    \Else
        \State $x^\prime_{ij} \gets 0$ \quad $\forall\, j$
    \EndIf
\EndFor
\State \Return $\mathbf{X}^\prime$
\end{algorithmic}
\end{algorithm}

\newpage
\section{Supplementary Material to Section 4.1}
\label{app:exp_setting}
\subsubsection{Metrics}
Performance is measured by AUROC, where the effect of each perturbation is given as $\Delta \text{AUROC} = \text{AUROC}_{\text{perturbed}} - \text{AUROC}_{\text{baseline}}$.
Statistical significance of performance degradation is assessed using the DeLong test~\cite{delong1988comparing}, applied directly to the aggregated predicted probabilities 
across all 5 folds. This yields approximately $n$ paired samples per test, one per dataset sample, providing substantially more statistical power than a fold-level test. Significance is assessed at $\alpha = 0.05$.

\section{Supplementary Material to Section 5}

\subsection{Informative vs.\ uninformative feature selection}
\label{app:feature_selection}

\begin{figure}[h!]
    \includegraphics[width=\linewidth]{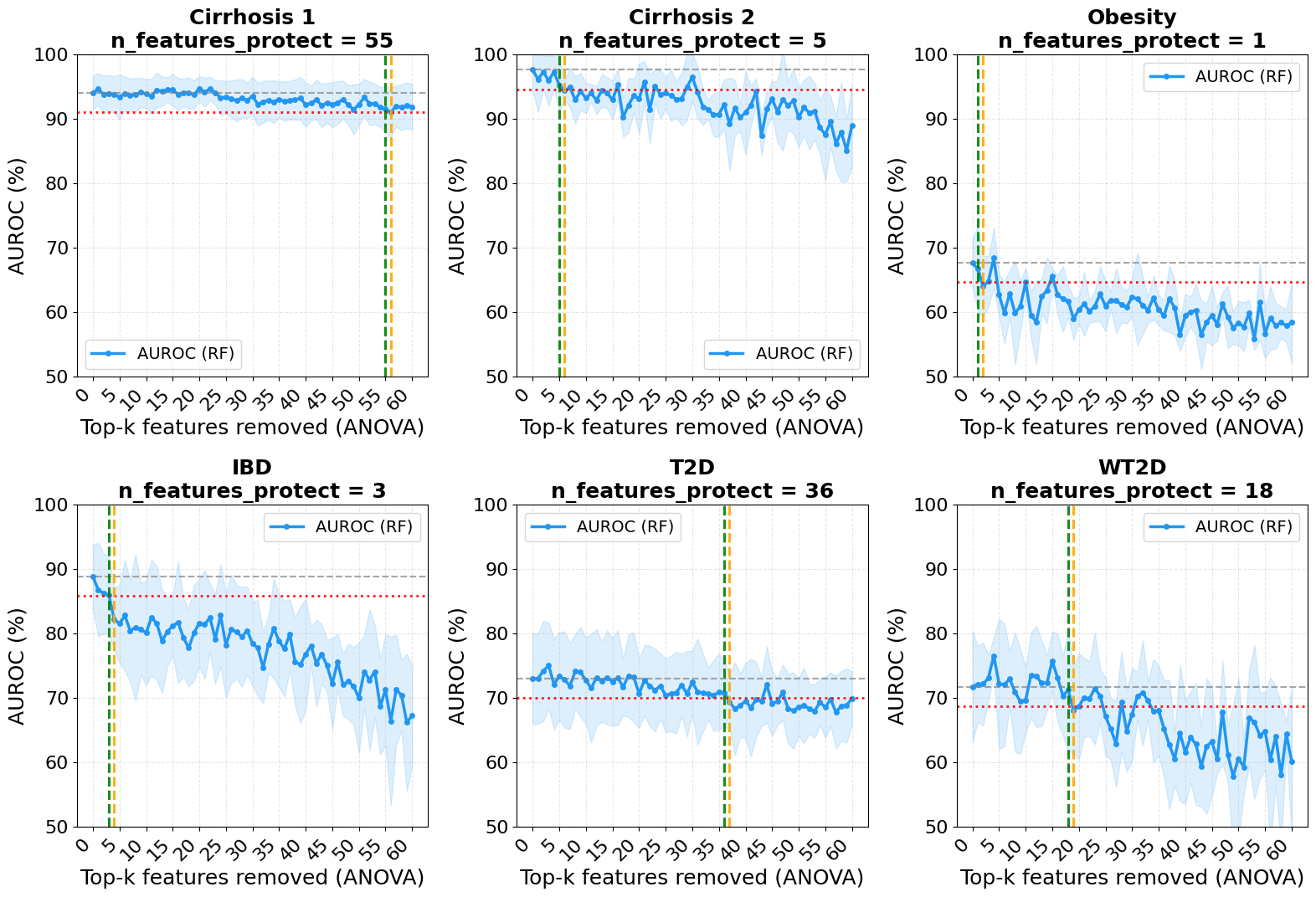}
    \caption{\textbf{AUROC vs top-k features removed} -- Features are ranked by ANOVA F-score, and a Random Forest is evaluated via 5-fold CV as top features are iteratively removed. The number of protected features is set just before AUROC drops by 3\%}.
    \label{fig:protected_features}
\end{figure}

\begin{table}[h]
\centering
\caption{\textbf{Number of informative selected features per dataset} -- $|\mathcal{F}_{I}|$.}
\begin{tabular}{l c c}
\toprule
\textbf{Dataset} & \textbf{AUROC} & \textbf{\#Protected} \\
\midrule
Cirrhosis 1 & 94.1 & 55 \\
Cirrhosis 2 & 97.6 & 5 \\
Obesity & 67.7 & 1 \\
IBD & 88.8 & 3 \\
T2D & 73.0 & 36 \\
WT2D & 71.2 & 18 \\
\bottomrule
\end{tabular}
\label{tab:protected_features}
\end{table}

\subsection{Semantic richness - robustness tradeoff}
\label{app:semantic}

\begin{table*}[t]
\centering
\caption{Baseline AUROC (mean $\pm$ std) across 5-fold CV of SOTA ML and TFM on Pasolli dataset\cite{pasolli2017accessible}.}
\label{tab:baseline_auroc}
\begin{tabular}{l|c|c|c|c|c|c}
\toprule
Model & Cirrhosis disc. & Cirrhosis val. & IBD & Obesity & T2D & WT2D \\
\midrule
$N$ (Control/Case) & 178 (83/95) & 54 (31/23) & 110 (85/25) & 253 (89/164) & 344 (174/170) & 96 (43/53) \\
\midrule
RF & 93.5 $\pm$ 3.3 & \textbf{95.9} $\pm$ 6.0 & 88.6 $\pm$ 6.4 & 65.0 $\pm$ 2.9 & 72.4 $\pm$ 7.3 & 66.2 $\pm$ 10.1 \\
XGBoost & 93.1 $\pm$ 3.2 & 88.6 $\pm$ 8.9 & 75.8 $\pm$ 10.3 & 63.6 $\pm$ 5.7 & 69.0 $\pm$ 7.4 & 68.3 $\pm$ 13.6 \\
\midrule
ContextTab & 93.7 $\pm$ 2.7 & 95.0 $\pm$ 6.7 & 90.3 $\pm$ 5.4 & 66.8 $\pm$ 3.6 & 75.4 $\pm$ 6.9 & 77.2 $\pm$ 9.8 \\
TabDPT & 87.7 $\pm$ 7.4 & 77.2 $\pm$ 10.7 & 80.2 $\pm$ 11.8 & 62.0 $\pm$ 7.9 & 69.4 $\pm$ 7.9 & 57.2 $\pm$ 11.7 \\
TabICL & \textbf{94.5} $\pm$ 2.9 & 92.1 $\pm$ 4.8 & 88.7 $\pm$ 7.1 & \textbf{70.4} $\pm$ 6.1 & \textbf{76.6} $\pm$ 7.0 & \textbf{81.9} $\pm$ 15.3 \\
TabPFN & 92.5 $\pm$ 2.8 & 94.5 $\pm$ 6.4 & \textbf{91.8} $\pm$ 5.7 & 63.7 $\pm$ 4.5 & 73.1 $\pm$ 7.7 & 74.3 $\pm$ 15.4 \\
\bottomrule
\end{tabular}
\end{table*}

The concordance analysis, together with the per-perturbation results, points to a broader pattern. \textit{TabICL} and \textit{ContextTab} co-degrade under zero inflation despite having different architectures. What they share is that both have learned realistic abundance distributions during pre-training. In \textit{TabICL}, the column-wise Set Transformer builds per-feature marginals that feed into CLS-based row embeddings~\cite{qu2025tabicl}, zeroing out values collapses these marginals. In \textit{ContextTab}, column-name embeddings tie each feature to a learned lexical representation of the corresponding taxon name; 
this semantic grounding becomes uninformative when the abundance values it depends on are replaced by zeros.

This points to a tradeoff between semantic richness and robustness. Models that have learned how to represent a feature, whether through distributional statistics or learned column embeddings, are also more sensitive to how that feature is corrupted.

\textit{TabPFN}, trained on synthetic data without domain-specific priors, largely avoids this vulnerability ($n_{\text{sig}} = 3$ under zero inflation). For practitioners, this means that the choice between semantically rich and distribution-agnostic models should depend on the expected data quality at deployment. When sparsity and compositional structure are stable across training and test environments, semantic models can leverage their richer representations. When these properties vary, as is common in multi-site microbiome studies with different sequencing protocols, a model with fewer distributional assumptions may be the safer choice.
\label{app:tables}
\clearpage

\end{document}